\documentclass[letterpaper]{article} % DO NOT CHANGE THIS
\usepackage{aaai2026}  % DO NOT CHANGE THIS
\usepackage{times}  % DO NOT CHANGE THIS
\usepackage{helvet}  % DO NOT CHANGE THIS
\usepackage{courier}  % DO NOT CHANGE THIS
\usepackage[hyphens]{url}  % DO NOT CHANGE THIS
\usepackage{graphicx} % DO NOT CHANGE THIS
\usepackage{natbib}  % DO NOT CHANGE THIS AND DO NOT ADD ANY OPTIONS TO IT
\usepackage{caption} % DO NOT CHANGE THIS AND DO NOT ADD ANY OPTIONS TO IT
\usepackage{algorithm}
\usepackage{algorithmic}

\usepackage{newfloat}
\usepackage{listings}
\DeclareCaptionStyle{ruled}{labelfont=normalfont,labelsep=colon,strut=off} % DO NOT CHANGE THIS
\floatstyle{ruled}
\newfloat{listing}{tb}{lst}{}
\floatname{listing}{Listing}
\nocopyright

\usepackage{fancyhdr}

\usepackage{amsmath}
\usepackage{tikz}

\title{\textsc{GraphVSSM}: Graph Variational State-Space Model \\for Probabilistic Spatiotemporal Inference \\of Dynamic Exposure and Vulnerability\\for Regional Disaster Resilience Assessment}
\author{
    Joshua Dimasaka\textsuperscript{\rm 1,\rm 2},
    Christian Gei{\ss}\textsuperscript{\rm 3, \rm 4},
    Emily So\textsuperscript{\rm 1, \rm 2}
}
\affiliations{
    \textsuperscript{\rm 1}Department of Architecture, University of Cambridge, United Kingdom\\
    \textsuperscript{\rm 2}Cambridge University Centre for Risk in the Built Environment, United Kingdom\\
    \textsuperscript{\rm 3}Earth Observation Center, German Aerospace Center, We{\ss}ling, Germany\\
    \textsuperscript{\rm 4}Institute of Geography, University of Bonn, Bonn, Germany\\
    jtd33@cam.ac.uk}

\begin{document}

\maketitle
\begin{tikzpicture}[remember picture,overlay]
  \node[
    anchor=south,
    yshift=56pt,                      % slightly above the bottom edge
    xshift=0pt,
    text width=\paperwidth,
    align=center,
    font=\small
  ] at (current page.south) {%
    \thepage{} -- Preprint -- Under Review%
  };
\end{tikzpicture}

\begin{abstract}
Regional disaster resilience quantifies the changing nature of physical risks to inform policy instruments ranging from local immediate recovery to international sustainable development. While many existing state-of-practice methods have greatly advanced the dynamic mapping of exposure and hazard, our understanding of large-scale physical vulnerability has remained static, costly, limited, region-specific, coarse-grained, overly aggregated, and inadequately calibrated. With the significant growth in the availability of time-series satellite imagery and derived products for exposure and hazard, we focus our work on the equally important yet challenging element of the risk equation: physical vulnerability. Given this unique problem, we leverage machine learning methods that flexibly capture spatial contextual relationships, limited temporal observations, and uncertainty in a unified probabilistic spatiotemporal inference framework. We therefore introduce Graph Variational State-Space Model (\textbf{\textsc{GraphVSSM}}), a novel modular spatiotemporal approach that uniquely integrates graph deep learning, state-space modeling, and variational inference using time-series data and prior expert belief systems in a weakly supervised or coarse-to-fine-grained manner. We present three major results: a city-wide demonstration in Quezon City, Philippines; an investigation of sudden changes in the cyclone-impacted coastal Khurushkul community (Bangladesh) and mudslide-affected Freetown (Sierra Leone); and an open geospatial dataset, \textbf{\textsc{METEOR 2.5D}}, that spatiotemporally enhances the existing global static dataset for 46 UN-recognized Least Developed Countries (as of 2020). Beyond advancing the practice of regional disaster resilience assessment and improving our understanding of global progress in disaster risk reduction, our method also offers a probabilistic deep learning approach, contributing to broader urban studies that require compositional data analysis in weakly supervised settings.
\end{abstract}

\begin{links}
    \link{Code}{https://github.com/riskaudit/GraphVSSM}
    \link{Quezon City (Philippines) Dataset}{https://doi.org/pzj2}
    \link{METEOR 2.5D Dataset, Part 1/2}{https://doi.org/pzq4}
    \link{METEOR 2.5D Dataset, Part 2/2}{https://doi.org/pzrd}
    \link{Khurushkul-Freetown Dataset}{https://doi.org/pzkw}
\end{links}

%%%%%%%%%%%%%%%%%%%%%%%%%%%%%% INTRODUCTION %%%%%%%%%%%%%%%%%%%%%%%%%%%%%%
\section{Introduction}

When a large earthquake, widespread flooding, raging wildfire, or any natural hazard strikes, we frequently hear the call for ``resilience'', which has unfortunately become increasingly ambiguous and overused in many recent global climate conversations \cite{bogardi2019intriguing, parker2020disaster}. In contrast, when designing buildings and critical infrastructure, quantifying physical (or asset) resilience follows a clearer framework as a dynamic risk, which is a temporal convolution of three major changing elements -- exposure, vulnerability, and hazard -- through time \cite{bruneau2003framework, applied2012seismic, almufti2013resilience}. Varying temporal scales of dynamic risk analysis serve different yet interconnected purposes, ranging from immediate recovery efforts \cite{comerio2006estimating} at daily-to-annual scales to development-oriented climate finance \cite{asian2017climate} at longer intervals, such as five-year periods. For example, low- and middle-income countries in the Caribbean, Pacific, and Africa have begun to benefit from modern international climate finance instruments, such as regional catastrophic risk pools \cite{ciullo2023increasing}. This underscores the importance of large-scale monitoring of changing patterns of exposure and vulnerability, amid intensifying climatic hazards, to better assess regional disaster resilience.

Despite recent advances in dynamic exposure and hazard modelling -- such as Google Open Buildings 2.5D Temporal \cite{sirko2023high}, DLR World Settlement Footprint Evolution \cite{marconcini2021understanding}, Global Human Settlement Layer multitemporal products \cite{pesaresi2024advances}, Microsoft Aurora \cite{bodnar2025foundation}, Google GraphCast \cite{lam2023learning}, and ECMWF AIFS \cite{lang2024aifs} -- progress in mapping the dynamic nature of physical vulnerability remains limited \cite{undrr2025gar}. Current approaches still rely heavily on the static information from the GEM Global Exposure Map and Vulnerability Model \cite{yepes2023global,martins2023global} and the METEOR project \cite{huyck2019meteor}. Although analytical Bayesian probabilistic approaches \cite{pittore2020variable,gomez2022towards} and various dasymmetric disaggregation techniques \cite{geiss2023benefits, dimasaka2024globalmappingexposurephysical, dimasaka2025deepc4} have enhanced spatial accuracy, the intricate subjectivity and coarse-grained representation of physical vulnerability as an ungeneralizable label across different construction practices indicate that solutions must be highly region-specific and consider the prevailing weakly supervised setup in both spatial and temporal domains. 

With the significant growth in the availability of time-series satellite imagery, its rich information allows us to recognize relevant patterns and tackle these challenges using machine learning methods that flexibly capture spatial relationships, temporal dynamics, and uncertainty in a unified framework \cite{rolf2024mission}. In particular, graph deep learning methods have enhanced our ability to model spatial relationships of irregularly structured data (e.g. connections between buildings) compared to conventional convolutional neural networks \cite{bronstein2021geometric, sheikh2025graph}. Additionally, deep state-space methods combine probabilistic sequential learning with prior knowledge specification and deep variational inference for uncertainty quantification (e.g., the changes in the likelihood of a highly vulnerable label given an existing coarse-grained belief system) \cite{girin2020dynamical, lin2024deep}. Together, these advanced capabilities open new opportunities to address the gap in modeling the interpretable regional dynamics of exposure and physical vulnerability towards a more fine-grained and accurate large-scale mapping while incorporating existing and valuable prior domain knowledge. 

In this paper, our key contributions are:

\begin{itemize}
    \item \textbf{\textsc{GraphVSSM}} or Graph Variational State-Space Model, a novel spatiotemporal method that integrates graph deep learning, state-space modeling, and variational inference using time-series satellite imagery and prior expert belief systems in a weakly supervised manner.
    \item A probabilistic graph-based deep learning framework that advances the purely analytical Bayesian updating approach of previous work \cite{pittore2020variable}, enabling the learning of complex relationships between building height, local spatial connectivity, and region-specific vulnerability labels.
    \item A city-scale demonstration of the assumed probability distributions for various indicators of regional exposure and physical vulnerability, using Quezon City in the Philippines as a case study.
    \item A high-resolution demonstration of yearly changes in physical vulnerability from 2016 to 2023 in the cyclone-impacted coastal Khurushkul community in Bangladesh (also known as \emph{``the world's largest climate refugee rehabilitation project''} \cite{khan4996026tropical}) and mudslide-affected Freetown in Sierra Leone in 2017.
    \item A novel open geospatial dataset, \textbf{\textsc{METEOR 2.5D}}, that spatially refines the existing METEOR dataset (i.e., from 450-meter to 90-meter scale) and significantly extends it with country-wide dynamic evolution of regional exposure and physical vulnerability for 46 UN-recognized Least Developed Countries as of 2020, provided at five-year intervals between 1975 and 2030.
\end{itemize}

%%%%%%%%%%%%%%%%%%%%%%%%%%%%%%%%% Related Work

\section{Related Work}

This section reviews related literature across three key topics that motivate our development of \textbf{\textsc{GraphVSSM}} for modeling the dynamic geospatial nature of physical vulnerability. We first examine current approaches and assumptions in analytical and statistical dynamic models used in disaster risk science, highlighting the distinct challenge of weak supervision. We then discuss how their limitations motivate the use of graph deep learning for efficient representation learning of building attributes. Next, we highlight the suitability of deep state-space models for handling short time series, which is another core aspect of our problem setting. Finally, we present the emerging concept of a graph state-space model and describe how extending it with variational inference supports a unified probabilistic framework for capturing uncertainty within disaster risk practice.

\subsection{Dynamic Mapping of Physical Vulnerability}

Along with the rapid growth of volunteered geographic information and satellite-derived products, many recent studies have developed a variety of analytical and probabilistic mapping approaches for dynamically updating the distribution of physical vulnerability across space and time \cite{cremen2022modelling}. Relevant to our scope of large-scale mapping, the pioneering work of \citeauthor{porter2014user} \shortcite{porter2014user} introduced the use of Beta probability distribution to perform Bayesian updating of the likelihood of a single building type using outcomes from field surveys. Building on categorical Bayesian inference \cite{agresti2005bayesian}, \citeauthor{pittore2020variable} \shortcite{pittore2020variable} extended this approach using Dirichlet and Multinomial probability distributions to account for the correlation among different building types. 

In contrast to these past studies that explicitly incorporate sparse in situ data, \citeauthor{lallemant2015modeling} \shortcite{lallemant2015modeling} applied a cellular automata approach and Markov chain modeling to dynamically update the physical vulnerability (e.g., incrementally expanding buildings), despite the lack of data calibration \cite{lallemant2017framework}. Differently, \citeauthor{calderon2022forecasting} \shortcite{calderon2022forecasting} proposed representing each location as a multi-agent system whose behavior depends on a geographically weighted regression model. As a more interpretable and manual yet costly approach, \citeauthor{schorlemmer2020global} \shortcite{schorlemmer2020global} synthesized large volumes of crowdsourced information, such as OpenStreetMap, using a rule-based approach. 

All of these past studies point toward the same prevailing challenge posed by the weak supervision setting, where we aim to refine coarse-grained vulnerability labels into finer spatiotemporal scales. Our study seeks to advance these state-of-practice methods by using deep learning to flexibly capture rich information from diverse datasets, including time-series satellite imagery, while also incorporating prior expert belief systems. This enables us to derive probabilistic representations that are better suited to disaster risk practice.

\subsection{Graph Deep Learning of Building Attributes}

Problems in weather, climate, and urban modeling that utilize satellite imagery have seen significant advances through the adoption of graph deep learning. The technique leverages contextual information and handles unstructured data, making it particularly well-suited for spatiotemporal graph-structured datasets \cite{zhao2024beyond}. 

For instance, \citeauthor{fill2024predicting} \shortcite{fill2024predicting} constructed distance-based subgraphs to predict residential building type across multiple countries, outperforming convolutional neural networks. Likewise, \citeauthor{xu2022building} \shortcite{xu2022building} and \citeauthor{lei2024predicting} \shortcite{lei2024predicting} demonstrated that graph deep learning achieved higher performance than support vector machines and random forests. \citeauthor{kong2024graph} \shortcite{kong2024graph} further emphasized its ability to capture correlations among different building types. For other unstructured data, \citeauthor{dimasaka2024enhancing} \shortcite{dimasaka2024enhancing} effectively linked sparse settlement areas with complex road networks at large scales.

Therefore, considering the sparsity of the building footprints in our problem setting, our study combines deep learning with
graph-structured representations to provide greater flexibility
in modeling diverse representations and attributes of the built environment.

\subsection{Deep State-Space Model for Short Sequences}

In time-series forecasting, real-world scenarios with limited historical observations -- particularly in spatiotemporal studies that rely on coarse temporal resolutions, such as yearly, five-year, or ten-year intervals -- motivate the use of simpler models to avoid the overfitting tendency of more complex ones \cite{lim2021time}. In our problem setting where publicly available moderate-resolution satellite imagery has recently been introduced to inform key global frameworks through 2030, the characteristics of temporal data in terms of information quantity, irregularity, and sequence length are crucial considerations in deciding between data-driven or deep forecasting models \cite{benidis2022deep}.

To tackle these challenges, \citeauthor{benidis2022deep} \shortcite{benidis2022deep} further noted that deep state-space models outperform purely autoregressive and discriminative models, particularly in low-data regimes \cite{de2020normalizing} and  when dealing with noisy or incomplete data (e.g., gaps in satellite imagery or sparsity of satellite-derived products). Specifically, \citeauthor{rangapuram2018deep} \shortcite{rangapuram2018deep} showed that hybridizing state-space models with deep learning facilitates an interpretable and generalizable framework by efficiently reducing the hypothesis space using input prior knowledge to the latent states  \cite{grover2015deep} and combining the flexibility of deep neural networks with the useful structural assumptions of traditional state-space models via latent state representations \cite{durbin2012time}. Hence, in our study, we adopt a deep state-space model because its explicit structural assumptions align well with estimating the likelihood of physical vulnerability classes based on building height patterns and prior knowledge from expert-informed census-derived products.

Building on the preceding section on graph deep learning, our study therefore advances deep state-space modeling by integrating graph neural networks, which capture more inductive relational biases in terms of neighborhood effects among buildings that share the same local contextual characteristics. While early formulations of graph state-space models have primarily used encoder-decoder architectures \cite{zambon2023graph, cini2025graph}, we expand this approach by incorporating variational learning to parameterize the latent space representations probabilistically \cite{blei2017variational}, leveraging the noisy signals of satellite imagery. This framing aligns well with the probabilistic treatment within disaster risk science, enabling our proposed approach to coherently quantify uncertainties across regional exposure, vulnerability, and hazard \cite{ward2020natural}.

%%%%%%%%%%%%%%%%%%%%%%%%%%%%%%%%% GraphVSSM

\section{Graph Variational State-Space Model}

\begin{figure}[t]
\centering
\includegraphics[width=1.0\columnwidth]{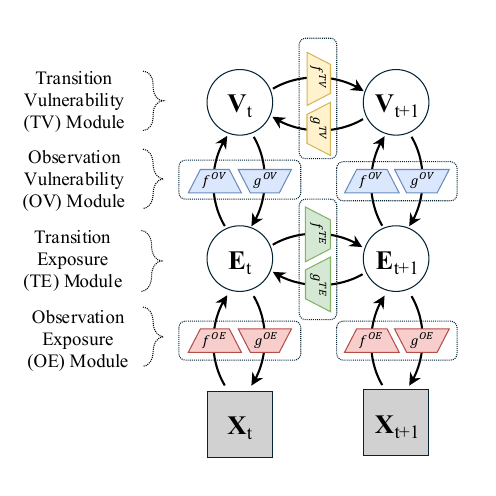} % Reduce the figure size so that it is slightly narrower than the column. Don't use precise values for figure width.This setup will avoid overfull boxes.
\caption{A schematic diagram of \textbf{\textsc{GraphVSSM}} consisting of four key modular components, where $\mathbf{X}$, $\mathbf{E}$, and $\mathbf{V}$ represent any independent explanatory variables (e.g., signals of satellite imagery and other auxiliary data such as proximity to road networks), exposure (e.g., probability of building presence and height), and vulnerability (e.g., probability of a particular building typology), respectively. Each module has a variational autoencoder with graph convolutional networks \cite{kipf2016semi} for encoder (i.e., $f \in \{{f^{\mathbf{OE}}}, {f^{\mathbf{TE}}} {f^{\mathbf{OV}}}, {f^{\mathbf{TV}}}\}$) and decoder (i.e., $g \in \{{g^{\mathbf{OE}}}, {g^{\mathbf{TE}}} {g^{\mathbf{OV}}}, {g^{\mathbf{TV}}}\}$), signifying relationships between and dynamics within $\mathbf{X}$, $\mathbf{E}$, and $\mathbf{V}$.}
\label{fig1}
\end{figure}

This section presents the formulation of \textbf{\textsc{GraphVSSM}}, as illustrated in Figure \ref{fig1}. We first discuss the probabilistic representations of our exposure and physical vulnerability, followed by how these representations are integrated into each modular component of our graph state-space model. Next, we present the supervised variational learning that incorporates input prior knowledge. Finally, we show how this approach benefits from the expressive representation of spatiotemporal graphs, including its features and assumptions behind our design of the adjacency matrices.

\subsection{Probabilistic Models}
\label{probModels}

We define the exposure $\mathbf{E}$ with two distinct probabilistic assumptions for the continuous and discrete variables of building height and presence, respectively. Upon initial empirical investigation of their signal distributions and considering the non-negative constraint, we assume building height ($\mathbf{E_{BH}}$) and presence ($\mathbf{E_{BP}}$) to follow lognormal and Bernoulli probability distributions, respectively, as:
\begin{equation}
    \ln \mathbf{E_{BH}} \in \{ \ln\mathbf{E_{BH}^{OE}}, \ln\mathbf{E_{BH}^{TE}} \} 
    \sim \mathcal{N}(\boldsymbol{\mu}_{\theta}(\mathbf{X}),\,\boldsymbol{\Sigma}_{\theta}(\mathbf{X}))
    \label{eq:XBH}
\end{equation}
\begin{equation}
    \mathbf{E_{BP}} \in \{ \mathbf{E_{BP}^{OE}}, \mathbf{E_{BP}^{TE}} \} \sim \mathrm{Bern}(\boldsymbol{p}_{\theta}(\mathbf{X}))
    \label{eq:XBP}
\end{equation}
\noindent where $\boldsymbol{\mu}_{\theta}$, $\boldsymbol{\Sigma}_{\theta}$, and $\boldsymbol{p}_{\theta}$ (i.e., via logit $\boldsymbol{\ell}_{\theta}$) are outcomes of our encoder networks, $f^{\mathbf{OE}}$ and $f^{\mathbf{TE}}$, with learnable parameters, $\theta$.

Moreover, following the analytical formulation of \citeauthor{pittore2020variable} \shortcite{pittore2020variable}, we also represent the vulnerability $\mathbf{V}$ as a categorical multinomial random variable:
\begin{equation}
    \mathbf{V} \in \{ \mathbf{V^{OV}}, \mathbf{V^{TV}} \} \sim \mathrm{Mult}(\boldsymbol{p}_{\theta}^{1}(\mathbf{X}), \ldots, \boldsymbol{p}_{\theta}^{K}(\mathbf{X}))
    \label{eq:XV}
\end{equation}
\noindent where $\boldsymbol{p}_{\theta}$ are outcomes of our encoder networks, $f^{\mathbf{OV}}$ and $f^{\mathbf{TV}}$, with learnable parameters $\theta$. 

In a given single pixel that can have varying spatial resolutions in our experiments, we want to determine the probability of observing $k^{th}$ building typology from all possible $K$ types. Unlike their approach that uses discrete counts of buildings \cite{pittore2020variable}, we improve this analytical approach to use the rasterized proportions of buildings, enabling the efficient use of geospatial data across large areal extents.

\subsection{State-Space Model Components}

For ease of demonstration, we sequentially train the modules in the following order: $\mathbf{OE}$, $\mathbf{TE}$, $\mathbf{OV}$, and $\mathbf{TV}$. This modularity provides flexibility, for cases where only the relationship between $\mathbf{E}$ and $\mathbf{V}$ is found to be more relevant than the well-studied task of predicting or forecasting $\mathbf{E}$. Nevertheless, our full training has still enabled us to establish and validate our probabilistic assumptions for $\mathbf{E}$ and $\mathbf{V}$. 

\subsubsection{$\mathbf{OE}$: Observation Exposure Module}

The encoder network ${f^{\mathbf{OE}}}$ inputs the $\mathbf{X}$ covariates and jointly outputs the probabilistic parameters $\boldsymbol{\mu}_{\theta}^{\mathbf{OE}}$, $\boldsymbol{\Sigma}_{\theta}^{\mathbf{OE}}$, and $\boldsymbol{p}_{\theta}^{\mathbf{OE}}$ for $\mathbf{E_{BH}^{OE}}$ and $\mathbf{E_{BP}^{OE}}$. The decoder network ${g^{\mathbf{OE}}}$ inputs the samples of $\mathbf{E_{BH}^{OE}}$ and $\mathbf{E_{BP}^{OE}}$ and outputs $\mathbf{\hat{X}}$, a reconstruction of our covariates.

\subsubsection{$\mathbf{TE}$: Transition Exposure Module}

The encoder network ${f^{\mathbf{TE}}}$ inputs the samples of $\mathbf{E_{BH}^{OE}}$ and $\mathbf{E_{BP}^{OE}}$ at time $t$, appended with the previous ${g^{\mathbf{TE}}}$-reconstructed covariates (i.e., not the original $\mathbf{X}$) for the same time step, $t$. 

Then, it jointly outputs a different set of probabilistic parameters $\boldsymbol{\mu}_{\theta}^{\mathbf{TE}}$, $\boldsymbol{\Sigma}_{\theta}^{\mathbf{TE}}$, and $\boldsymbol{p}_{\theta}^{\mathbf{TE}}$ for $\mathbf{E_{BH}^{TE}}$ and $\mathbf{E_{BP}^{TE}}$ for the future time step, $t+1$. 

Similarly, the decoder network ${g^{\mathbf{TE}}}$ inputs the samples of $\mathbf{E_{BH}^{TE}}$ and $\mathbf{E_{BP}^{TE}}$ from a future time step, $t+1$, and backward reconstructs our inputs to ${f^{\mathbf{TE}}}$, for the preceding step, $t$. 

After training the ${f^{\mathbf{TE}}}$ and ${g^{\mathbf{TE}}}$, we sample $\mathbf{E_{BP}^{TE}}$ for the entire time horizon to identify the reduced set of pixels with a high likelihood of building presence (i.e., $\boldsymbol{\mu}_{\theta}^{\mathbf{TE}} \geq 0.5$), instead of using the full map. This new set of pixels concludes the probabilistic modeling of $\mathbf{E}$ and introduces a different set of adjacency matrices for the succeeding probabilistic modeling of $\mathbf{V}$.

\subsubsection{$\mathbf{OV}$: Observation Vulnerability Module}

The encoder network ${f^{\mathbf{OV}}}$ inputs the samples of $\mathbf{E_{BH}^{TE}}$ and the previously ${g^{\mathbf{TE}}}$-reconstructed covariates. Then, it outputs the probabilistic $K$ parameters $\boldsymbol{p}_{\theta}^{\mathbf{OV}}$, corresponding to the $K$ types of physical vulnerability. The decoder network ${g^{\mathbf{OV}}}$ inputs the samples from this multinomial probability distribution and outputs a further reconstruction of our covariates.

\subsubsection{$\mathbf{TV}$: Transition Vulnerability Module}

The encoder network ${f^{\mathbf{TV}}}$ inputs the samples of $\mathbf{V^{OV}}$ at time $t$, appended with the previous ${g^{\mathbf{OV}}}$-reconstructed covariates for the same time step, $t$. Then, it outputs a different set of probabilistic $K$ parameters $\boldsymbol{p}_{\theta}^{\mathbf{TV}}$ for the next time step, $t+1$. Similarly, the decoder network ${g^{\mathbf{TV}}}$ inputs the samples of $\mathbf{V^{TV}}$ from a future step, $t+1$, and then backward reconstructs our inputs to ${f^{\mathbf{TV}}}$, for the preceding step, $t$. 

\subsection{Variational Learning}

\subsubsection{Encoder Reparameterization Tricks} To enable the variational nature, we implement the following reparameterization trick for the stochastic samplings of our building height variable using:
\begin{equation}
    \mathbf{E_{BH}^{*}} = e^{\boldsymbol{\mu}_{\theta}+\boldsymbol{\Sigma}_{\theta}^{1/2}\boldsymbol{\epsilon^{*}}}, \quad \boldsymbol{\epsilon^{*}} \sim  \mathcal{N}(0,1)
    \label{eq:ept_lognormal}
\end{equation}

Similarly, the variables for building presence and proportion of vulnerability types follow the same categorical parameterization trick via Gumbel-Softmax distribution for a continuous, differentiable approximation, proposed by \citeauthor{jang2016categorical} \shortcite{jang2016categorical}. In symbols,
\begin{equation}
    \{ \mathbf{E_{BP}^{*}}, \mathbf{V^{*}}\} = \frac{
        e^{(\boldsymbol{\ell_{\theta,k}} + \boldsymbol{g^{*}_k})/\tau}
    }{
        \sum_{j=1}^{K} e^{(\boldsymbol{\ell_{\theta,j}} +  \boldsymbol{g^{*}_j})/\tau}
    }
    \quad \text{for } k = 1, \dots, K
    \label{eq:ept_categorical}
\end{equation}

\noindent where $K=2$ for binary $\mathbf{E_{BP}}$ and $K\geq2$ for multiclass $\mathbf{V}$. $\tau$ is our scalar temperature input upon checking the shape of prior distribution across $K$ classes, and $\boldsymbol{g^{*}}$ is sampled as:
\begin{equation}
    \boldsymbol{g^{*}} = -\log(-\log(\boldsymbol{u^{*}})), \quad \boldsymbol{u^{*}} \sim  \mathrm{Uniform}(0,1)
    \label{eq:ept_categorical_g}
\end{equation}

In addition, as introduced in Section \ref{probModels}, $\boldsymbol{\ell_{\theta}}$ are logits, which are direct outputs of our encoder networks. To determine the corresponding probabilistic parameter $\boldsymbol{p}_{\theta}$ for the $k^{th}$ class or building typology for our Bernoulli or multinomial probabilistic distribution, respectively, we can also use the softmax operator but without the stochastic sampling part and the scalar temperature input, as:
\begin{equation}
    \boldsymbol{p}^{k}_{\theta} = \frac{
        e^{\boldsymbol{\ell_{\theta,k}}}
    }{
        \sum_{j=1}^{K} e^{\boldsymbol{\ell_{\theta,j}}}
    }
\end{equation}

\subsubsection{Local Pruning for $K$ Vulnerability Types}

Following a similar strategy for efficient training from multi-hazard impact studies \cite{xu2022seismic}, the first of our two ways of incorporating prior knowledge information is through local pruning of our multinomial probabilistic distribution. For example, even though a region can have $K$ vulnerability types, some areas can have a varying number of types. To illustrate, a more rural or more urbanized area might have only fewer types, while the rural-urban interface area can have more diverse types up to $K$ types. In our experiment, the logit is simply set to a small value (e.g., $0.001$) if the $k^{th}$ building typology is known a priori to be non-existent.

\subsubsection{Loss Functions}

We train every module using the sum of reconstruction loss, $\mathcal{L}^{\text{rec}}$, and Kullback-Leibler divergence loss, $\mathcal{L}^{\text{KL}}$.
The training of all decoder networks, $g$, is more straightforward than that of our encoder networks, $f$. 

On one hand, our decoder networks minimize $\mathcal{L}^{\text{rec}}$ in terms of a weighted mean-squared error.
\begin{equation}
    \mathcal{L}^{\text{rec}} = \frac{1}{N} \sum_{i=1}^{N} w^{x}_i \, (x_i - \hat{x}_i)^2
\end{equation}

\noindent where, for $N$ data points, $x \in \mathbf{X}$ and $\hat{x} \in \mathbf{\hat{X}}$. $w^{x}$ is a scalar shape-dependent weight with values depending on their Euclidean distances from the defining characteristic points of the empirical distribution of $\mathbf{X}$. For instance, since $\mathbf{X}$ seems to follow a normal distribution upon investigation, we assign higher weights to points located at the modal peak and at both extreme ends; otherwise, the weights for other points are linearly interpolated.

On the other hand, the encoder networks minimize $\mathcal{L}^{\text{KL}}$ and use a similar weight assignment approach. For $\mathbf{E_{BH}}$, we take the average in such a way that, for $i^{th}$ point,
\begin{equation}
    \begin{split}
    \mathcal{L}_{\mathbf{BH},i}^{\text{KL}} = 
    \frac{1}{2} \Bigg(
        & \frac{(\mu_{\theta,i} - \mu_{0,i})^2}{\sigma_{0,i}^2} w^{\mu_{0}}_i \\
        & + \Bigg(\frac{\sigma_{\theta,i}^2}{\sigma_{0,i}^2}
          - \ln \Big( \frac{\sigma_{\theta,i}^2}{\sigma_{0,i}^2} \Big)
          - 1\Bigg) w^{\sigma_{0}}_i
    \Bigg)
    \end{split}
\end{equation}
\noindent where $\mu_{\theta,i} \in \boldsymbol{\mu}_{\theta}$ and $\sigma_{\theta,i} \in \boldsymbol{\sigma}_{\theta}$. Similarly, $\mu_{0}$, $\sigma_{0}$, $w^{\mu_{0}}$, and $w^{\sigma_{0}}$ are from the prior distribution of building height.

For $\mathbf{E_{BP}}$, 
\begin{equation}
    \mathcal{L}_{\mathbf{BP},i}^{\text{KL}}
    = \Bigg(p_{\theta,i} \log \frac{p_{\theta,i}}{p_{0,i}} + (1 - p_{\theta,i}) \log \frac{1 - p_{\theta,i}}{1 - p_{0,i}}\Bigg) w^{p_{0}}_i
\end{equation}
\noindent where $p_{\theta,i} \in \boldsymbol{p}_{\theta}$. Similarly, $p_{0}$ and $w^{p_{0}}$ are from the prior distribution of building presence.

For $\mathbf{V}$ where $i^{th}$ point has $K$ possible types,
\begin{equation}
    \mathcal{L}_{\mathbf{V},i}^{\text{KL}}
    = \Bigg(\sum_{k=1}^K p_{\theta,i,k} \log \frac{p_{\theta,i,k}}{p_{0,i,k}} \Bigg) w^{p_{0}}_i
\end{equation}
\noindent where $p_{\theta,i} \in \boldsymbol{p}_{\theta}$. $p_{0}$ and $w^{p_{0}}$ also come from the prior distribution of vulnerability types. Specifically, $w^{p_{0}}$ assigns higher importance to the most likely class via masking.

In $\mathbf{V}$, our initial findings suggest that the effects of weak supervision combined with many possible classes posed difficulties in learning diverse classifications. Hence, we added a supervised cross-entropy loss from a semi-supervised variational learning solution \cite{kingma2014semi}. 
\begin{equation}
    \mathcal{L}_{\mathbf{V},i}^{\text{CE}}
    = \Bigg(\sum_{k=1}^K p_{0,i,k} \log(p_{\theta,i,k})\Bigg) w^{p_{0}}_i
\end{equation}

\subsection{Spatiotemporal Graph}

Whether at city or country scale, we divide the region into multiple square tiles and split them into training, testing, and validation sets with a balanced number of $\mathbf{V}$ classes. Using the following spatiotemporal graph representations, we train $\mathbf{OE}$, $\mathbf{TE}$, $\mathbf{OV}$, and $\mathbf{TV}$ modules with graph convolutional networks \cite{kipf2016semi}.

For each tile, we create an undirected exposure graph $G^{E}_{t}=(N^{E}_{t}, A^{E}_{t}, X^{E}_{t})$ at time step $t$. $N^{E}_{t}$ is the set of nodes that represent all pixels in the square tile, $A^{E}_{t}$ is the grid-based adjacency matrix or connectivity information between these nodes from all eight directions, and $X^{E}_{t}$ is the feature covariates such as the signals and derived indices from satellite imagery. Through time, while the $A^{E}_{t}$ remains unchanged, the values of $X^{E}_{t}$ may vary to indicate changes in building presence and height.

To train the $\mathbf{OV}$ and $\mathbf{TV}$ modules, we then reduce the $A^{E}_{t}$ if the sampled values of $\mathbf{E_{BP}^{TE}}$ exceeded $0.5$. In this case, the $A^{E}_{t}$ becomes $A^{V}_{t}$, which identifies $N^{V}_{t}$ and also changes through time. Our $X^{V}_{t}$ contains the sampled values of $\mathbf{E_{BH}^{TE}}$ and, as mentioned previously, the previously ${g^{\mathbf{TE}}}$-reconstructed covariates, thereby creating the undirected vulnerability graph $G^{V}_{t}$.

\subsection{Evaluation}

To facilitate a comparable evaluation with previous studies on dynamic vulnerability modelling \cite{pittore2020variable}, we calculate the Aitchison distance, $d_A$, which quantifies the difference between the compositions of two different models (i.e., prior $\boldsymbol{p}_{0}$ and ours $\boldsymbol{p}_{\theta}$) \cite{aitchison1982statistical}.
\begin{equation}
    d_A = \sqrt{ \frac{1}{2K} \sum_{i=1}^K \sum_{j=1}^K 
    \left[ \ln\left( \frac{\boldsymbol{p}_{0,i}}{\boldsymbol{p}_{0,j}} \right) - \ln\left( \frac{\boldsymbol{p}_{\theta,i}}{\boldsymbol{p}_{\theta,j}} \right) \right]^2 }
\end{equation}

%%%%%%%%%%%%%%%%%%%%%%%%%%%%%%%%% Experimental Setup

\section{Experimental Setup}

We demonstrate the application of \textbf{\textsc{GraphVSSM}} through three distinct case studies, each with varying spatiotemporal resolutions. First, we implement full sequential training of all modules for Quezon City, Philippines, at an annual 10-meter resolution, conducted in close collaboration with the disaster risk managers. Next, to account for significant changes caused by recent disasters, we showcase the benefits of modularity by training only the $\mathbf{OV}$ module for the cyclone-impacted coastal Khurushkul community (Bangladesh) and the mudslide-affected Freetown (Sierra Leone), both at an annual 50-centimeter resolution. Finally, to illustrate the scalability at a coarser continental level, we train the $\mathbf{OV}$ module for 46 UN-recognized Least Developed Countries as of 2020, using a five-year interval at a 90-meter resolution. Further details are in Appendix \ref{appendix:experimentalsetup}.

%%%%%%%%%%%%%%%%%%%%%%%%%%%%%%%%% Results and Discussion

\section{Results and Discussion}

\subsection{Changing Physical Vulnerability Composition}

\begin{figure*}[t]
    \centering
    \includegraphics[width=7in,height=5in]{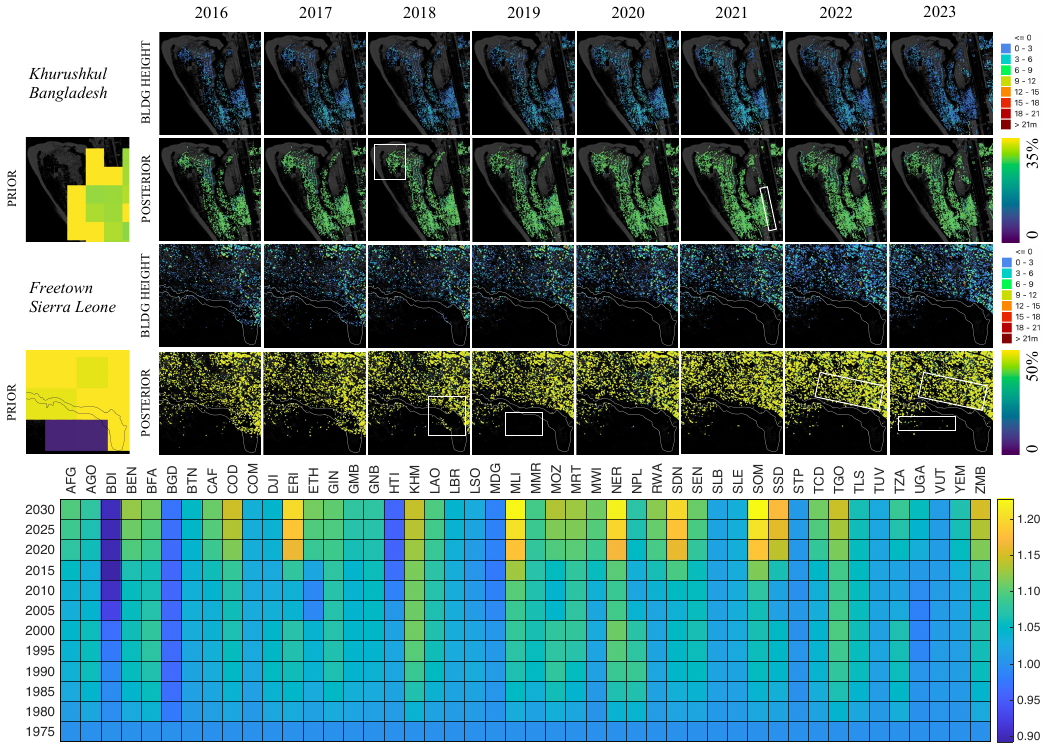}
    \caption{\emph{Top}: Annual changes in building presence and height (Google Open Buildings 2.5D Temporal) and the corresponding METEOR prior and $\mathbf{OV}$ posterior probabilities for the informal settlement (INF) and unreinforced concrete block masonry (UCB) classes in cyclone-impacted Khurushkul (Bangladesh) and mudslide-affected Freetown (Sierra Leone), respectively. The full-resolution figures are available in the supplementary material. \emph{Bottom}: Change in mean Aitchison distance between METEOR prior and $\mathbf{OV}$ posterior probabilities (from GHSL multitemporal products) of vulnerability classes across 46 Least Developed Countries (as of 2020), calculated as the ratio relative to the 1975 baseline (T/1975), at five-year intervals, 1975–2030.}
    \label{fig2}
\end{figure*}

In Figure \ref{fig2}, we illustrate, through two different time scales, the changes in the composition of physical vulnerability using our trained $\mathbf{OV}$ module with time-series satellite-derived products of building exposure. 

In Khurushkul, Bangladesh, our findings show that prior to the disaster, our posterior maps for 2016 and 2017 indicated an increasing density of informal constructions around the coastal community. However, this pattern lessened in 2018 following the disaster and began to show signs of scattered rebuilding from 2019 onward, with a notable expansion near the northwest part of the map (i.e., close to the water body). The 2021 posterior map also captured the significant displacement caused by the construction of the airport. Since the building height and geometry of the new airport facilities have different characteristics from that of the surrounding dominant informal settlement patterns, the trained $\mathbf{OV}$ module recognized these differently, assigning lower probabilities of being classified as informal constructions. This result underscores the influence of local contextual information (i.e., how the houses are sized and irregularly arranged) as captured by the graph neural network, on the probability of a particular building typology.

\begin{figure*}[t]
    \centering
    \includegraphics[width=7in,height=1.3in]{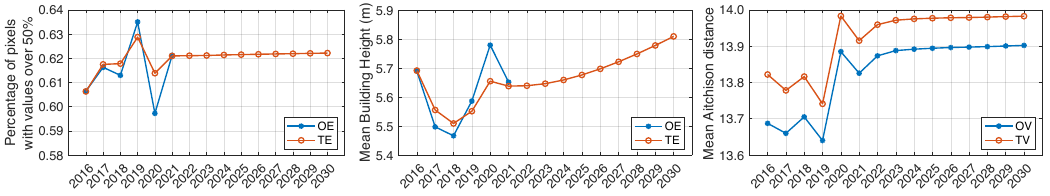} 
    \caption{Annual changes in $\mathbf{OE}$/$\mathbf{TE}$ posterior parameters for $\mathbf{E_{BP}}$ \emph{(left)} and $\mathbf{E_{BH}}$ \emph{(middle)}, and the mean Aitchison distance between prior and $\mathbf{OV}$/$\mathbf{TV}$ posterior probabilities of $\mathbf{V}$ classes \emph{(right)} in Quezon City, Philippines.}
    \label{fig3}
\end{figure*}

In Freetown, Sierra Leone, although the mudslide caused the affected area to be uninhabitable, our posterior maps revealed a post-disaster community-level behavior in which a densification of the same weak vulnerability class (unreinforced concrete block masonry) occurred around the perimeter and, in some instances, even within the previously devastated area. Since both cases show the persistence of highly vulnerable building types in areas with elevated climate and disaster risks, our evidence collectively suggest the critical importance of regularly monitoring the spatial and temporal distribution of physical vulnerability to better inform a holistic community-level disaster resilience planning.

On the bottom half of Figure \ref{fig2}, the change in mean Aitchison distance supports our prior expert belief systems regarding the compositions of vulnerability classes. Particularly, our resulting heatmap inflection points around 2000-2010 (i.e., shifting from decreasing to increasing) suggest that the METEOR data likely reflects vulnerability patterns during that decade. Between 1975 and 2000, the positive rate of change corresponds to the influence of a low built-up area characterized by a more homogeneous composition of vulnerability classes. After 2010, the rate of change accelerates beyond that of the 1975–2000 period, indicative of the anticipated urban expansion and substantial changes in the heterogeneous composition of physical vulnerability classes.

\subsection{Uncovering Spatiotemporal Dynamics}

Examining the derived posterior parameters across our four modules of the graph state-space model reveals insights into the regional spatiotemporal dynamics of building exposure and physical vulnerability. As shown in Figure \ref{fig3}, during the observable 2016-2021 period, the trends of posterior parameters for $\mathbf{OE}$, $\mathbf{TE}$, $\mathbf{OV}$, and $\mathbf{TV}$ remained similar but began to diverge after 2021, implying the learned slow temporal behavior in building exposure and physical vulnerability. The annual spatiotemporal maps are available in the supplementary material.

\subsection{Learning Interpretable Probabilistic Latents}

\begin{figure}[t!]
    \centering
    \includegraphics[width=3.3in,height=3.72in]{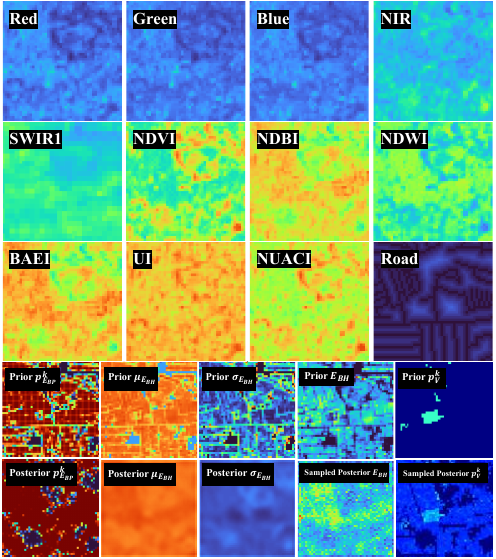} 
    \caption{\emph{Top twelve}: Explanatory covariates and derived indices from satellite imagery. \emph{Bottom ten}: Comparisons between prior and $\mathbf{OE}$/$\mathbf{OV}$ posterior parameters.}
    \label{fig4}
\end{figure}

Representing latent variables as parameters of assumed probability distributions has become feasible through the use of multiple explanatory covariates and derived indices from satellite imagery, as shown in Figure \ref{fig4}. Our results further demonstrate that incorporating prior knowledge into the Kullback-Leibler divergence and cross-entropy losses to address weak multi-class supervision allows the framework to balance supervised learning with deep Bayesian coarse-to-fine-grained updating. This yields interpretable posterior parameters and enables the reconstruction of covariates using reparameterization techniques for both continuous and categorical distributions.

%%%%%%%%%%%%%%%%%%%%%%%%%%%%%%%%% Conclusion and Future Work 

\section{Conclusion and Future Work}

While machine learning and time-series satellite imagery are increasingly used to understand climate and disaster risks -- especially for dynamic exposure and hazard mapping -- our work tackles the equally important yet uniquely challenging aspect of modeling physical vulnerability at large spatiotemporal scales. Our contribution demonstrates a probabilistic deep learning approach, contributing to broader urban studies that require compositional data analysis in weakly supervised or coarse-to-fine-grained settings. For future study, we recommend the following next steps:

\begin{itemize}

    \item In training the graph state-space model, while we understand that this may become computationally expensive, simultaneous, instead of sequential, learning can provide insight into a unified modelling framework.
    \item In vulnerability modelling, the use of vector-based representation of buildings can result in an interesting modification of the multinomial probability distribution that uses discrete counts, instead of rasterized proportions.
    \item Calibrating the $\mathbf{TV}$ module with monitored information on the changes of physical vulnerability (e.g., retrofit, demolition, or disaster damage) can improve the overall performance and flexibility with new observations.
    
\end{itemize}

\section*{Acknowledgments}

This work is funded by the UKRI Centre for Doctoral Training in Application of Artificial Intelligence to the study of Environmental Risks (AI4ER) (EP/S022961/1). We are thankful for the following individuals and organizations: the building exposure and physical vulnerability data from Quezon City Disaster Risk Reduction and Management Office (Ms. Maria Bianca Perez); the valuable practical guidance from Earthquakes and Megacities Initiative (Dr. Fouad Bendimerad and Dr. Renan Tanhueco); Charles Huyck for sharing technical limitations of the METEOR products; and Dr. Robert Muir-Wood for proposing the need for spatiotemporal disaster risk quantification auditing and technical suggestions on the relevance of case studies.

\bibliography{aaai2026}

\begin{thebibliography}{63}
\providecommand{\natexlab}[1]{#1}

\bibitem[{{ADB}(2017)}]{asian2017climate}
{ADB}. 2017.
\newblock \emph{Climate Change Operational Framework, 2017-2030: Enhanced Actions for Low Greenhouse Gas Emissions and Climate-resilient Development}.
\newblock Asian Development Bank.

\bibitem[{Agresti and Hitchcock(2005)}]{agresti2005bayesian}
Agresti, A.; and Hitchcock, D.~B. 2005.
\newblock Bayesian inference for categorical data analysis.
\newblock \emph{Statistical Methods and Applications}, 14: 297--330.

\bibitem[{Aitchison(1982)}]{aitchison1982statistical}
Aitchison, J. 1982.
\newblock The statistical analysis of compositional data.
\newblock \emph{Journal of the Royal Statistical Society: Series B (Methodological)}, 44(2): 139--160.

\bibitem[{Allen et~al.(2014)Allen, Ryu, Bautista, Bautista, Narag, Sevilla, Melosantos, Papiona, and Bonita}]{allen2014enhancing}
Allen, T.; Ryu, H.; Bautista, B.; Bautista, M.~L.; Narag, I.; Sevilla, W.; Melosantos, M.~L.; Papiona, K.; and Bonita, J. 2014.
\newblock Enhancing risk analysis capacities for flood, tropical cyclone severe wind and earthquake for the greater metro manila area.
\newblock \emph{Philippine Institute of Volcanology and Seismology, Geoscience Australia}.

\bibitem[{Almufti and Willford(2013)}]{almufti2013resilience}
Almufti, I.; and Willford, M. 2013.
\newblock The resilience-based earthquake design initiative (REDiTM) rating system.
\newblock \emph{Arup Co}.

\bibitem[{Benidis et~al.(2022)Benidis, Rangapuram, Flunkert, Wang, Maddix, Turkmen, Gasthaus, Bohlke-Schneider, Salinas, Stella et~al.}]{benidis2022deep}
Benidis, K.; Rangapuram, S.~S.; Flunkert, V.; Wang, Y.; Maddix, D.; Turkmen, C.; Gasthaus, J.; Bohlke-Schneider, M.; Salinas, D.; Stella, L.; et~al. 2022.
\newblock Deep learning for time series forecasting: Tutorial and literature survey.
\newblock \emph{ACM Computing Surveys}, 55(6): 1--36.

\bibitem[{Blei, Kucukelbir, and McAuliffe(2017)}]{blei2017variational}
Blei, D.~M.; Kucukelbir, A.; and McAuliffe, J.~D. 2017.
\newblock Variational inference: A review for statisticians.
\newblock \emph{Journal of the American statistical Association}, 112(518): 859--877.

\bibitem[{Bodnar et~al.(2025)Bodnar, Bruinsma, Lucic, Stanley, Allen, Brandstetter, Garvan, Riechert, Weyn, Dong et~al.}]{bodnar2025foundation}
Bodnar, C.; Bruinsma, W.~P.; Lucic, A.; Stanley, M.; Allen, A.; Brandstetter, J.; Garvan, P.; Riechert, M.; Weyn, J.~A.; Dong, H.; et~al. 2025.
\newblock A foundation model for the Earth system.
\newblock \emph{Nature}, 641(8065): 1180--1187.

\bibitem[{Bogardi and Fekete(2019)}]{bogardi2019intriguing}
Bogardi, J.~J.; and Fekete, A. 2019.
\newblock From intriguing concept (s) towards an overused buzzword: is it time for a requiem for resilience?
\newblock \emph{Resilience and Vulnerability: Conceptual}, 1996.

\bibitem[{Bronstein et~al.(2021)Bronstein, Bruna, Cohen, and Veli{\v{c}}kovi{\'c}}]{bronstein2021geometric}
Bronstein, M.~M.; Bruna, J.; Cohen, T.; and Veli{\v{c}}kovi{\'c}, P. 2021.
\newblock Geometric deep learning: Grids, groups, graphs, geodesics, and gauges.
\newblock \emph{arXiv preprint arXiv:2104.13478}.

\bibitem[{Bruneau et~al.(2003)Bruneau, Chang, Eguchi, Lee, O'Rourke, Reinhorn, Shinozuka, Tierney, Wallace, and Von~Winterfeldt}]{bruneau2003framework}
Bruneau, M.; Chang, S.~E.; Eguchi, R.~T.; Lee, G.~C.; O'Rourke, T.~D.; Reinhorn, A.~M.; Shinozuka, M.; Tierney, K.; Wallace, W.~A.; and Von~Winterfeldt, D. 2003.
\newblock A framework to quantitatively assess and enhance the seismic resilience of communities.
\newblock \emph{Earthquake spectra}, 19(4): 733--752.

\bibitem[{Calderon and Silva(2022)}]{calderon2022forecasting}
Calderon, A.; and Silva, V. 2022.
\newblock Forecasting seismic risk within the context of the Sendai framework: An application to the Dominican Republic.
\newblock \emph{International Journal of Disaster Risk Reduction}, 82: 103364.

\bibitem[{Cini et~al.(2025)Cini, Marisca, Zambon, and Alippi}]{cini2025graph}
Cini, A.; Marisca, I.; Zambon, D.; and Alippi, C. 2025.
\newblock Graph Deep Learning for Time Series Forecasting.
\newblock \emph{ACM Comput. Surv.}

\bibitem[{Ciullo et~al.(2023)Ciullo, Strobl, Meiler, Martius, and Bresch}]{ciullo2023increasing}
Ciullo, A.; Strobl, E.; Meiler, S.; Martius, O.; and Bresch, D.~N. 2023.
\newblock Increasing countries’ financial resilience through global catastrophe risk pooling.
\newblock \emph{Nature Communications}, 14(1): 922.

\bibitem[{Comerio(2006)}]{comerio2006estimating}
Comerio, M.~C. 2006.
\newblock Estimating downtime in loss modeling.
\newblock \emph{Earthquake Spectra}, 22(2): 349--365.

\bibitem[{{Copernicus Sentinel data}(2025{\natexlab{a}})}]{sentinel2}
{Copernicus Sentinel data}. 2025{\natexlab{a}}.
\newblock {Harmonized Sentinel-2 MSI: MultiSpectral Instrument, Level-2A}.
\newblock \url{https://developers.google.com/earth-engine/datasets/catalog/COPERNICUS_S2_SR_HARMONIZED}.
\newblock Accessed: 2025-02-01.

\bibitem[{{Copernicus Sentinel data}(2025{\natexlab{b}})}]{sentinel2cloud}
{Copernicus Sentinel data}. 2025{\natexlab{b}}.
\newblock {Sentinel-2: Cloud Probability}.
\newblock \url{https://developers.google.com/earth-engine/datasets/catalog/COPERNICUS_S2_CLOUD_PROBABILITY}.
\newblock Accessed: 2025-02-01.

\bibitem[{Cremen, Galasso, and McCloskey(2022)}]{cremen2022modelling}
Cremen, G.; Galasso, C.; and McCloskey, J. 2022.
\newblock Modelling and quantifying tomorrow's risks from natural hazards.
\newblock \emph{Science of The Total Environment}, 817: 152552.

\bibitem[{de~B{\'e}zenac et~al.(2020)de~B{\'e}zenac, Rangapuram, Benidis, Bohlke-Schneider, Kurle, Stella, Hasson, Gallinari, and Januschowski}]{de2020normalizing}
de~B{\'e}zenac, E.; Rangapuram, S.~S.; Benidis, K.; Bohlke-Schneider, M.; Kurle, R.; Stella, L.; Hasson, H.; Gallinari, P.; and Januschowski, T. 2020.
\newblock Normalizing kalman filters for multivariate time series analysis.
\newblock \emph{Advances in Neural Information Processing Systems}, 33: 2995--3007.

\bibitem[{Dimasaka, Geiß, and So(2024)}]{dimasaka2024globalmappingexposurephysical}
Dimasaka, J.; Geiß, C.; and So, E. 2024.
\newblock Global Mapping of Exposure and Physical Vulnerability Dynamics in Least Developed Countries using Remote Sensing and Machine Learning.
\newblock arXiv:2404.01748.

\bibitem[{Dimasaka, Geiß, and So(2025)}]{dimasaka2025deepc4}
Dimasaka, J.; Geiß, C.; and So, E. 2025.
\newblock DeepC4: Deep Conditional Census-Constrained Clustering for Large-scale Multitask Spatial Disaggregation of Urban Morphology.
\newblock arXiv:2507.22554.

\bibitem[{Dimasaka, Selvakumaran, and Marinoni(2024)}]{dimasaka2024enhancing}
Dimasaka, J.; Selvakumaran, S.; and Marinoni, A. 2024.
\newblock Enhancing assessment of direct and indirect exposure of settlement-transportation systems to mass movements by intergraph representation learning.
\newblock \emph{Environmental Research Letters}, 19(11): 114055.

\bibitem[{Durbin and Koopman(2012)}]{durbin2012time}
Durbin, J.; and Koopman, S.~J. 2012.
\newblock \emph{Time series analysis by state space methods}.
\newblock Oxford University Press (UK).

\bibitem[{{ECHO}(2017)}]{echo2017}
{ECHO}. 2017.
\newblock {Sierra Leone | Floods and Mudslides - DG ECHO Daily Map | 16/08/2017}.
\newblock \url{https://reliefweb.int/map/sierra-leone/sierra-leone-floods-and-mudslides-dg-echo-daily-map-16082017}.

\bibitem[{{EMI}(2022)}]{emi2022}
{EMI}. 2022.
\newblock Risk Profile Atlas (RPA) and Hazard, Vulnerability and Risk Assessment (HVRA) of Quezon City Government, Philippines.
\newblock Technical report, Earthquakes and Megacities Initiative and Quezon City Government.

\bibitem[{{FEMA}(2012)}]{applied2012seismic}
{FEMA}. 2012.
\newblock \emph{Seismic performance assessment of buildings}.
\newblock Federal Emergency Management Agency.

\bibitem[{Fill, Eichelbeck, and Ebner(2024)}]{fill2024predicting}
Fill, J.; Eichelbeck, M.; and Ebner, M. 2024.
\newblock Predicting building types and functions at transnational scale.
\newblock \emph{arXiv preprint arXiv:2409.09692}.

\bibitem[{Gei{\ss} et~al.(2023)Gei{\ss}, Priesmeier, Aravena~Pelizari, Soto~Calderon, Schoepfer, Riedlinger, Villar~Vega, Santa~Mar{\'\i}a, G{\'o}mez~Zapata, Pittore et~al.}]{geiss2023benefits}
Gei{\ss}, C.; Priesmeier, P.; Aravena~Pelizari, P.; Soto~Calderon, A.~R.; Schoepfer, E.; Riedlinger, T.; Villar~Vega, M.; Santa~Mar{\'\i}a, H.; G{\'o}mez~Zapata, J.~C.; Pittore, M.; et~al. 2023.
\newblock Benefits of global earth observation missions for disaggregation of exposure data and earthquake loss modeling: evidence from Santiago de Chile.
\newblock \emph{Natural Hazards}, 119(2): 779--804.

\bibitem[{Girin et~al.(2020)Girin, Leglaive, Bie, Diard, Hueber, and Alameda-Pineda}]{girin2020dynamical}
Girin, L.; Leglaive, S.; Bie, X.; Diard, J.; Hueber, T.; and Alameda-Pineda, X. 2020.
\newblock Dynamical variational autoencoders: A comprehensive review.
\newblock \emph{arXiv preprint arXiv:2008.12595}.

\bibitem[{G{\'o}mez~Zapata et~al.(2022)G{\'o}mez~Zapata, Zafrir, Pittore, and Merino}]{gomez2022towards}
G{\'o}mez~Zapata, J.~C.; Zafrir, R.; Pittore, M.; and Merino, Y. 2022.
\newblock Towards a sensitivity analysis in seismic risk with probabilistic building exposure models: an application in Valparaiso, Chile using ancillary open-source data and parametric ground motions.
\newblock \emph{ISPRS International Journal of Geo-Information}, 11(2): 113.

\bibitem[{Grover, Kapoor, and Horvitz(2015)}]{grover2015deep}
Grover, A.; Kapoor, A.; and Horvitz, E. 2015.
\newblock A deep hybrid model for weather forecasting.
\newblock In \emph{Proceedings of the 21th ACM SIGKDD international conference on knowledge discovery and data mining}, 379--386.

\bibitem[{Huyck et~al.(2019)Huyck, Hu, Amyx, Esquivias, Huyck, and Eguchi}]{huyck2019meteor}
Huyck, C.; Hu, Z.; Amyx, P.; Esquivias, G.; Huyck, M.; and Eguchi, M. 2019.
\newblock METEOR: exposure data classification, metadata population and confidence assessment. Report M3. 2/P.
\newblock Technical Report M3. 2/P, {British Geological Survey}.

\bibitem[{Jang, Gu, and Poole(2016)}]{jang2016categorical}
Jang, E.; Gu, S.; and Poole, B. 2016.
\newblock Categorical reparameterization with gumbel-softmax.
\newblock \emph{arXiv preprint arXiv:1611.01144}.

\bibitem[{Khan et~al.(2024)Khan, Farid, Sojib, Islam, Gogon, and Farid}]{khan4996026tropical}
Khan, M.; Farid, Z.~I.; Sojib, M. T.~H.; Islam, M.~A.; Gogon, M. I.~R.; and Farid, S.~I. 2024.
\newblock How a Tropical Super-Cyclone Triggered an Intergenerational Vulnerability: A Case Study on a Cyclone Displaced Community in Cox's Bazar.
\newblock \emph{Available at SSRN 4996026}.

\bibitem[{Kingma et~al.(2014)Kingma, Mohamed, Jimenez~Rezende, and Welling}]{kingma2014semi}
Kingma, D.~P.; Mohamed, S.; Jimenez~Rezende, D.; and Welling, M. 2014.
\newblock Semi-supervised learning with deep generative models.
\newblock \emph{Advances in neural information processing systems}, 27.

\bibitem[{Kipf and Welling(2016)}]{kipf2016semi}
Kipf, T.~N.; and Welling, M. 2016.
\newblock Semi-supervised classification with graph convolutional networks.
\newblock \emph{arXiv preprint arXiv:1609.02907}.

\bibitem[{Kong et~al.(2024)Kong, Ai, Zou, Yan, and Yang}]{kong2024graph}
Kong, B.; Ai, T.; Zou, X.; Yan, X.; and Yang, M. 2024.
\newblock A graph-based neural network approach to integrate multi-source data for urban building function classification.
\newblock \emph{Computers, Environment and Urban Systems}, 110: 102094.

\bibitem[{Lallemant(2015)}]{lallemant2015modeling}
Lallemant, D. 2015.
\newblock \emph{Modeling the future disaster risk of cities to envision paths towards their future resilience}.
\newblock Stanford University.

\bibitem[{Lallemant et~al.(2017)Lallemant, Burton, Ceferino, Bullock, and Kiremidjian}]{lallemant2017framework}
Lallemant, D.; Burton, H.; Ceferino, L.; Bullock, Z.; and Kiremidjian, A. 2017.
\newblock A framework and case study for earthquake vulnerability assessment of incrementally expanding buildings.
\newblock \emph{Earthquake spectra}, 33(4): 1369--1384.

\bibitem[{Lam et~al.(2023)Lam, Sanchez-Gonzalez, Willson, Wirnsberger, Fortunato, Alet, Ravuri, Ewalds, Eaton-Rosen, Hu et~al.}]{lam2023learning}
Lam, R.; Sanchez-Gonzalez, A.; Willson, M.; Wirnsberger, P.; Fortunato, M.; Alet, F.; Ravuri, S.; Ewalds, T.; Eaton-Rosen, Z.; Hu, W.; et~al. 2023.
\newblock Learning skillful medium-range global weather forecasting.
\newblock \emph{Science}, 382(6677): 1416--1421.

\bibitem[{Lang et~al.(2024)Lang, Alexe, Chantry, Dramsch, Pinault, Raoult, Clare, Lessig, Maier-Gerber, Magnusson et~al.}]{lang2024aifs}
Lang, S.; Alexe, M.; Chantry, M.; Dramsch, J.; Pinault, F.; Raoult, B.; Clare, M.~C.; Lessig, C.; Maier-Gerber, M.; Magnusson, L.; et~al. 2024.
\newblock AIFS--ECMWF's data-driven forecasting system.
\newblock \emph{arXiv preprint arXiv:2406.01465}.

\bibitem[{Lei et~al.(2024)Lei, Liu, Milojevic-Dupont, and Biljecki}]{lei2024predicting}
Lei, B.; Liu, P.; Milojevic-Dupont, N.; and Biljecki, F. 2024.
\newblock Predicting building characteristics at urban scale using graph neural networks and street-level context.
\newblock \emph{Computers, Environment and Urban Systems}, 111: 102129.

\bibitem[{Lim and Zohren(2021)}]{lim2021time}
Lim, B.; and Zohren, S. 2021.
\newblock Time-series forecasting with deep learning.
\newblock \emph{Philosophical Transactions: Mathematical, Physical and Engineering Sciences}, 379(2194): 1--14.

\bibitem[{Lin and Michailidis(2024)}]{lin2024deep}
Lin, J.; and Michailidis, G. 2024.
\newblock Deep Learning-based Approaches for State Space Models: A Selective Review.
\newblock \emph{arXiv preprint arXiv:2412.11211}.

\bibitem[{Marconcini et~al.(2021)Marconcini, Metz-Marconcini, Esch, and Gorelick}]{marconcini2021understanding}
Marconcini, M.; Metz-Marconcini, A.; Esch, T.; and Gorelick, N. 2021.
\newblock Understanding current trends in global urbanisation-the world settlement footprint suite.
\newblock \emph{GI\_Forum}, 9(1): 33--38.

\bibitem[{Martins and Silva(2023)}]{martins2023global}
Martins, L.; and Silva, V. 2023.
\newblock {Global Vulnerability Model of the GEM Foundation}.
\newblock https://doi.org/10.5281/zenodo.8391743.

\bibitem[{Parker(2020)}]{parker2020disaster}
Parker, D.~J. 2020.
\newblock Disaster resilience--a challenged science.

\bibitem[{Pesaresi et~al.(2024)Pesaresi, Schiavina, Politis, Freire, Krasnodebska, Uhl, Carioli, Corbane, Dijkstra, Florio et~al.}]{pesaresi2024advances}
Pesaresi, M.; Schiavina, M.; Politis, P.; Freire, S.; Krasnodebska, K.; Uhl, J.~H.; Carioli, A.; Corbane, C.; Dijkstra, L.; Florio, P.; et~al. 2024.
\newblock Advances on the Global Human Settlement Layer by joint assessment of Earth Observation and population survey data.
\newblock \emph{International Journal of Digital Earth}, 17(1): 2390454.

\bibitem[{Pittore, Haas, and Silva(2020)}]{pittore2020variable}
Pittore, M.; Haas, M.; and Silva, V. 2020.
\newblock Variable resolution probabilistic modeling of residential exposure and vulnerability for risk applications.
\newblock \emph{Earthquake Spectra}, 36(1\_suppl): 321--344.

\bibitem[{Porter et~al.(2014)Porter, Hu, Huyck, and Bevington}]{porter2014user}
Porter, K.; Hu, Z.; Huyck, C.; and Bevington, J. 2014.
\newblock User guide: Field sampling strategies for estimating building inventories.
\newblock \emph{GEM Foundation}.

\bibitem[{Rangapuram et~al.(2018)Rangapuram, Seeger, Gasthaus, Stella, Wang, and Januschowski}]{rangapuram2018deep}
Rangapuram, S.~S.; Seeger, M.~W.; Gasthaus, J.; Stella, L.; Wang, Y.; and Januschowski, T. 2018.
\newblock Deep state space models for time series forecasting.
\newblock \emph{Advances in neural information processing systems}, 31.

\bibitem[{Rolf et~al.(2024)Rolf, Klemmer, Robinson, and Kerner}]{rolf2024mission}
Rolf, E.; Klemmer, K.; Robinson, C.; and Kerner, H. 2024.
\newblock Mission Critical--Satellite Data is a Distinct Modality in Machine Learning.
\newblock \emph{arXiv preprint arXiv:2402.01444}.

\bibitem[{Schorlemmer et~al.(2020)Schorlemmer, Beutin, Cotton, Garcia~Ospina, Hirata, Ma, Nievas, Prehn, and Wyss}]{schorlemmer2020global}
Schorlemmer, D.; Beutin, T.; Cotton, F.; Garcia~Ospina, N.; Hirata, N.; Ma, K.-F.; Nievas, C.; Prehn, K.; and Wyss, M. 2020.
\newblock Global dynamic exposure and the OpenBuildingMap-a big-data and crowd-sourcing approach to exposure modeling.
\newblock In \emph{EGU General Assembly Conference Abstracts}, 18920.

\bibitem[{Sheikh and Saha(2025)}]{sheikh2025graph}
Sheikh, N.; and Saha, S. 2025.
\newblock Graph neural networks for multi-sensor Earth observation.
\newblock \emph{Deep Learning for Multi-Sensor Earth Observation}, 211--230.

\bibitem[{Sirko et~al.(2023)Sirko, Brempong, Marcos, Annkah, Korme, Hassen, Sapkota, Shekel, Diack, Nevo et~al.}]{sirko2023high}
Sirko, W.; Brempong, E.~A.; Marcos, J.~T.; Annkah, A.; Korme, A.; Hassen, M.~A.; Sapkota, K.; Shekel, T.; Diack, A.; Nevo, S.; et~al. 2023.
\newblock High-resolution building and road detection from sentinel-2.
\newblock \emph{arXiv preprint arXiv:2310.11622}.

\bibitem[{{UNDRR}(2025)}]{undrr2025gar}
{UNDRR}. 2025.
\newblock {Global Assessment Report on Disaster Risk Reduction 2025: Resilience Pays: Financing and Investing for our Future}.
\newblock Technical report, {United Nations Office for Disaster Risk Reduction, Geneva}.

\bibitem[{{UNITAR-UNOSAT}(2017)}]{unitarunosat2017}
{UNITAR-UNOSAT}. 2017.
\newblock {Damage Assessment in Khurushkul Union, Cox's Bazar District, Bangladesh}.
\newblock \url{https://unosat.org/products/2501}.

\bibitem[{Ward et~al.(2020)Ward, Blauhut, Bloemendaal, Daniell, de~Ruiter, Duncan, Emberson, Jenkins, Kirschbaum, Kunz et~al.}]{ward2020natural}
Ward, P.~J.; Blauhut, V.; Bloemendaal, N.; Daniell, J.~E.; de~Ruiter, M.~C.; Duncan, M.~J.; Emberson, R.; Jenkins, S.~F.; Kirschbaum, D.; Kunz, M.; et~al. 2020.
\newblock Natural hazard risk assessments at the global scale.
\newblock \emph{Natural Hazards and Earth System Sciences}, 20(4): 1069--1096.

\bibitem[{Xu et~al.(2022{\natexlab{a}})Xu, Dimasaka, Wald, and Noh}]{xu2022seismic}
Xu, S.; Dimasaka, J.; Wald, D.~J.; and Noh, H.~Y. 2022{\natexlab{a}}.
\newblock Seismic multi-hazard and impact estimation via causal inference from satellite imagery.
\newblock \emph{Nature Communications}, 13(1): 7793.

\bibitem[{Xu et~al.(2022{\natexlab{b}})Xu, He, Xie, Xie, Luo, and Xie}]{xu2022building}
Xu, Y.; He, Z.; Xie, X.; Xie, Z.; Luo, J.; and Xie, H. 2022{\natexlab{b}}.
\newblock Building function classification in Nanjing, China, using deep learning.
\newblock \emph{Transactions in GIS}, 26(5): 2145--2165.

\bibitem[{Yepes-Estrada et~al.(2023)Yepes-Estrada, Calderon, Costa, Crowley, Dabbeek, Hoyos, Martins, Paul, Rao, and Silva}]{yepes2023global}
Yepes-Estrada, C.; Calderon, A.; Costa, C.; Crowley, H.; Dabbeek, J.; Hoyos, M.~C.; Martins, L.; Paul, N.; Rao, A.; and Silva, V. 2023.
\newblock Global building exposure model for earthquake risk assessment.
\newblock \emph{Earthquake Spectra}, 39(4): 2212--2235.

\bibitem[{Zambon et~al.(2023)Zambon, Cini, Livi, and Alippi}]{zambon2023graph}
Zambon, D.; Cini, A.; Livi, L.; and Alippi, C. 2023.
\newblock Graph state-space models.
\newblock \emph{arXiv preprint arXiv:2301.01741}.

\bibitem[{Zhao et~al.(2024)Zhao, Chen, Xiong, Shi, Saha, and Zhu}]{zhao2024beyond}
Zhao, S.; Chen, Z.; Xiong, Z.; Shi, Y.; Saha, S.; and Zhu, X.~X. 2024.
\newblock Beyond Grid Data: Exploring graph neural networks for Earth observation.
\newblock \emph{IEEE Geoscience and Remote Sensing Magazine}.

\end{thebibliography}
\clearpage

\appendix

\section{Further Details on Experimental Setup}
\label{appendix:experimentalsetup}

\subsection{Case Study 1: Quezon City, Philippines}

In 2022, the government of Quezon City conducted its climate and disaster risk assessment using an improved geospatial building exposure and physical vulnerability database \cite{emi2022}, combining another prior study \cite{allen2014enhancing} and a high-resolution digital elevation map. In our work, we enhanced this baseline further using the aggregated, cloud-filtered signals of annual 10-meter maps of publicly available Sentinel 2 multispectral imagery \cite{sentinel2, sentinel2cloud}, proximity to road networks, and temporal building height data from Google Open Buildings 2.5D Temporal \cite{sirko2023high} to infer the likelihood of each vulnerability class at finer-grained spatial resolution with identified temporal attributes. Consequently, relevant to key local decision makers, this case study demonstrates a jurisdiction-wide spatiotemporal updating or auditing of the existing building exposure and physical vulnerability database, aided by advanced machine learning techniques and satellite imagery. 

This case study has 19 vulnerability classes:
\begin{enumerate}
    \item W1: Wood Frame with Area $\geq$ 500 square meters
    \item W2: Wood Frame with Area $\leq$ 500 square meters
    \item W3: Bamboo
    \item N: Makeshift
    \item C1: Reinforced Concrete Moment Frame
    \item C2: Reinforced Concrete Shear Walls
    \item C4: Reinforced Concrete Moment Frame \& Shear Walls
    \item PC2: Precast Frame
    \item CHB: Concrete Hollow Blocks
    \item URA: Adobe
    \item URM: Brick
    \item RM1: Flexible Diaphragm
    \item RM2: Rigid Diaphragm
    \item MWS: Partial Masonry, Wood, and/or Metal
    \item CWS: Partial Reinforced Concrete, Wood, and/or Metal
    \item PC2: Precast Frame
    \item S1: Steel Moment Frame
    \item S2: Steel Braced Frame
    \item S3: Steel Light Metal
\end{enumerate}

This case study is a result of an ongoing collaboration with Quezon City Disaster Risk Reduction and Management Office (QCDRRMO) and Earthquakes and Megacities Initiative (EMI). To ensure the usefulness of our outputs for disaster risk managers, we used the official city data as part of our inputs and also are not publicly available due to confidential information. However, publicly available alternatives such as crowdsourced information can be used, but these alternatives can have less quality of validation and accuracy.

\subsection{Case Study 2: Cyclone-Impacted Coastal Khurushkul (Bangladesh) and Mudslide- Affected Freetown (Sierra Leone)}

Investigating post-disaster areas with poor socioeconomic capacity such as the cyclone-impacted coastal Khurushkul community in Bangladesh and mudslide-affected Freetown in Sierra Leone is important in monitoring the effectiveness of policies and international aid towards immediate recovery and risk reduction. In both cases, we trained an $\mathbf{OV}$ module using Google Open Buildings 2.5D Temporal \cite{sirko2023high} to learn inductive relationships with the most probable annual vulnerability class from 2016 to 2023.

In May 2017, Cyclone Mora damaged over 700 houses in Khurushkul Union, Cox's Bazar District \cite{unitarunosat2017}. Despite being known as  \emph{``the world's largest climate refugee rehabilitation project''}, relocation efforts remain inadequate in addressing the recurring and intergenerational vulnerability \cite{khan4996026tropical}. Similarly, in August 2017, heavy rainfall triggered widespread mudslides and landslides that wiped out over 300 houses, displaced 6200, and killed over 400 with more than 600 missing in Freetown in Sierra Leone \cite{echo2017}. Thus, our work examines whether these devastated communities and their surrounding areas have effectively reduced their physical risks (i.e., moving away from the hazardous zones) or still face heightened risk from such climate hazards (i.e.,  any signs of building construction with weak physical vulnerability).

The Khurushkul case study has 9 vulnerability classes.

\begin{enumerate}
    \item C3L: Nonductile reinforced concrete frame with masonry infill walls low-rise
    \item C3M: Nonductile reinforced concrete frame with masonry infill walls mid-rise
    \item INF: Informal constructions
    \item M: Mud walls
    \item RS: Rubble stone (field stone) masonry  
    \item S: Steel
    \item UFB: Unreinforced fired brick masonry   
    \item W3: Wood light unbraced post and beam frame
    \item W5: Wattle and Daub (Walls with bamboo/light timber log/reed mesh and post)
\end{enumerate}

The Freetown case study has 7 vulnerability classes.

\begin{enumerate}
    \item A: Adobe blocks (unbaked sundried mud block) walls
    \item INF: Informal constructions
    \item RS: Rubble stone (field stone) masonry  
    \item UCB: Concrete block unreinforced masonry with lime or cement mortar
    \item UFB: Unreinforced fired brick masonry  
    \item W: Wood
    \item W5: Wattle and Daub (Walls with bamboo/light timber log/reed mesh and post)
\end{enumerate}

\subsection{Case Study 3: UN Least Developed Countries}

Relevant to large-scale disaster risk assessment across multiple countries by international organizations \cite{undrr2025gar}, we trained a country-specific $\mathbf{OV}$ module using DLR World Settlement Footprint Evolution \cite{marconcini2021understanding} and Global Human Settlement Layer multitemporal products \cite{pesaresi2024advances}. This demonstration extends the existing METEOR dataset for 46 UN-recognized Least Developed Countries as of 2020, resulting in a five-fold improvement in spatial resolution (i.e., from 450-meter to 90-meter scale) and adding valuable temporal attributes at five-year intervals between 1975 and 2030.

This case study has 27 vulnerability classes, wherein every country has a varying number of available vulnerability classes \cite{dimasaka2024globalmappingexposurephysical}.

\begin{enumerate}
    \item A: Adobe blocks (unbaked sundried mud block) walls
    \item C: Reinforced concrete
    \item C3L: Nonductile reinforced concrete frame with masonry infill walls low-rise
    \item C3M: Nonductile reinforced concrete frame with masonry infill walls mid-rise
    \item C3H: Nonductile reinforced concrete frame with masonry infill walls high-rise
    \item DS: Rectangular cut-stone masonry block
    \item INF: Informal constructions
    \item M: Mud walls  
    \item RE: Rammed Earth/Pneumatically impacted stabilized earth
    \item RM: Reinforced masonry
    \item RS: Rubble stone (field stone) masonry  
    \item RS1: Local field stones dry stacked (no mortar) with timber floors, earth, or metal roof
    \item RS2: Local field stones with mud mortar
    \item RS3: Local field stones with lime mortar
    \item S: Steel
    \item S1L: Steel moment frame low-rise
    \item S1M: Steel moment frame mid-rise     
    \item S3: Steel light frame  
    \item S5: Steel frame with unreinforced masonry infill walls  
    \item UCB: Concrete block unreinforced masonry with lime or cement mortar
    \item UFB: Unreinforced fired brick masonry  
    \item UFB1: Unreinforced brick masonry in mud mortar without timber posts  
    \item W: Wood
    \item W1: Wood stud-wall frame with plywood/gypsum board sheathing
    \item W2: Wood frame, heavy members (with area  5000 sq. ft.)    
    \item W3: Wood light unbraced post and beam frame
    \item W5: Wattle and Daub (Walls with bamboo/light timber log/reed mesh and post)
\end{enumerate}

\subsection{Computing Infrastructure}
We performed all experiments using a MacBook Pro (Apple M3 Max) with 48GB memory. Fortunately, our experiments did not need to use GPU. Due to ease of implementation and our familiarity, we used the deep learning and mapping toolboxes of MATLAB. However, other software libraries and frameworks can be used to reproduce our results.

\end{document}